\documentclass[sigconf]{acmart}
\AtBeginDocument{%
  \providecommand\BibTeX{{%
    \normalfont B\kern-0.5em{\scshape i\kern-0.25em b}\kern-0.8em\TeX}}}

\renewcommand\footnotetextcopyrightpermission[1]{} % removes footnote with conference information in first column
\setcopyright{none}

\acmConference[ACM, SIGKDD]{}
\acmYear{2025}
\acmBooktitle{ACM, SIGKDD}

\usepackage{algorithmicx,algorithm}
\usepackage{algpseudocode}
\usepackage{algorithm,algpseudocode}
\usepackage{makecell,booktabs}
\usepackage{multirow}
\usepackage{multicol}  
\usepackage{bm}
\usepackage{siunitx}
\usepackage{xcolor}
\usepackage{enumitem}
\usepackage{subfigure}
\usepackage{xspace}
\usepackage{cleveref}
\usepackage{etoolbox}
\usepackage{pdfcomment}
\usepackage{amsmath}
\usepackage[normalem]{ulem}
\usepackage{marvosym}   % corresponding envelop
\usepackage{soul}
\usepackage{pifont}
\newcommand{\multirowoffset}{-0.5\dimexpr \aboverulesep + \belowrulesep + \cmidrulewidth}
\newcommand{\cmark}{\text{\ding{51}}}
\newcommand{\xmark}{\text{\ding{55}}}

\def\model{OmniTraj\xspace}
\usepackage{color, xspace}

\begin{document}
\title{Learning Generalized and Flexible Trajectory Models from Omni-Semantic Supervision
}

\author{Yuanshao Zhu$^{1,2,3~\ast}$,~James Jianqiao Yu$^{4 ~\dag}$,~Xiangyu Zhao$^{2~\dag}$, Xiao Han$^2$, \\ Qidong Liu$^2$, Xuetao Wei$^1$, Yuxuan Liang$^{3~\dag}$}
\thanks{$^{\ast}$ Work done during the internship at HKUST(GZ)\\
$^{\dag}$ Corresponding authors.}
\affiliation{
  \institution{$^1$Southern University of Science and Technology,~~$^2$City University of Hong Kong}
   \country{}
  \address{}
} 
\affiliation{
  \institution{$^3$The Hong Kong University of Science and Technology (Guangzhou),~~$^4$Harbin Institute of Technology, Shenzhen}
   \country{}
  \address{}
  \text{yuanshao@ieee.org,~jqyu@ieee.org,~xianzhao@cityu.edu.hk,~hahahenha@gmail.com}\\
\text{liuqidong@stu.xjtu.edu.cn,~weixt@sustech.edu.cn,~~yuxliang@outlook.com}
}

\renewcommand{\shortauthors}{Yuanshao Zhu et al.}

\begin{abstract}
The widespread adoption of mobile devices and data collection technologies has led to an exponential increase in trajectory data, presenting significant challenges in spatio-temporal data mining, particularly for efficient and accurate trajectory retrieval. 
However, existing methods for trajectory retrieval face notable limitations, including inefficiencies in large-scale data, lack of support for condition-based queries, and reliance on trajectory similarity measures. 
To address the above challenges, we propose \textbf{OmniTraj}, a generalized and flexible omni-semantic trajectory retrieval framework that integrates four complementary modalities or semantics---raw trajectories, topology, road segments, and regions---into a unified system. 
Unlike traditional approaches that are limited to computing and processing trajectories as a single modality, OmniTraj designs dedicated encoders for each modality, which are embedded and fused into a shared representation space.
This design enables OmniTraj to support accurate and flexible queries based on any individual modality or combination thereof, overcoming the rigidity of traditional similarity-based methods.
Extensive experiments on two real-world datasets demonstrate the effectiveness of OmniTraj in handling large-scale data, providing flexible, multi-modality queries, and supporting downstream tasks and applications.
\end{abstract}

\maketitle

\section{Introduction}

The widespread availability of GPS-enabled devices and urban sensing infrastructure has resulted in the generation of vast amounts of spatio-temporal trajectory data, capturing the movement patterns of vehicles, pedestrians, and public urban services \cite{wang2021survey,chen2024deep,deng2022multi,zhu2024unitraj}. 
As a fundamental building block for smart city applications, trajectory retrieval enables semantically rich insight into mobility behaviors, ranging from traffic congestion prediction and emergency route planning to location-based recommendations \cite{su2020survey,chang2023trajectory,fang2022spatio}. 
For instance, urban planners might analyze trajectories passing through specific road segments to optimize traffic light scheduling, while logistics companies could query delivery routes that intersect designated regions to evaluate operational efficiency \cite{su2013calibrating,guo2020context,deng2011trajectory,deng2024parsimony}.

\begin{figure}[!t]
    \includegraphics[width=1\linewidth]{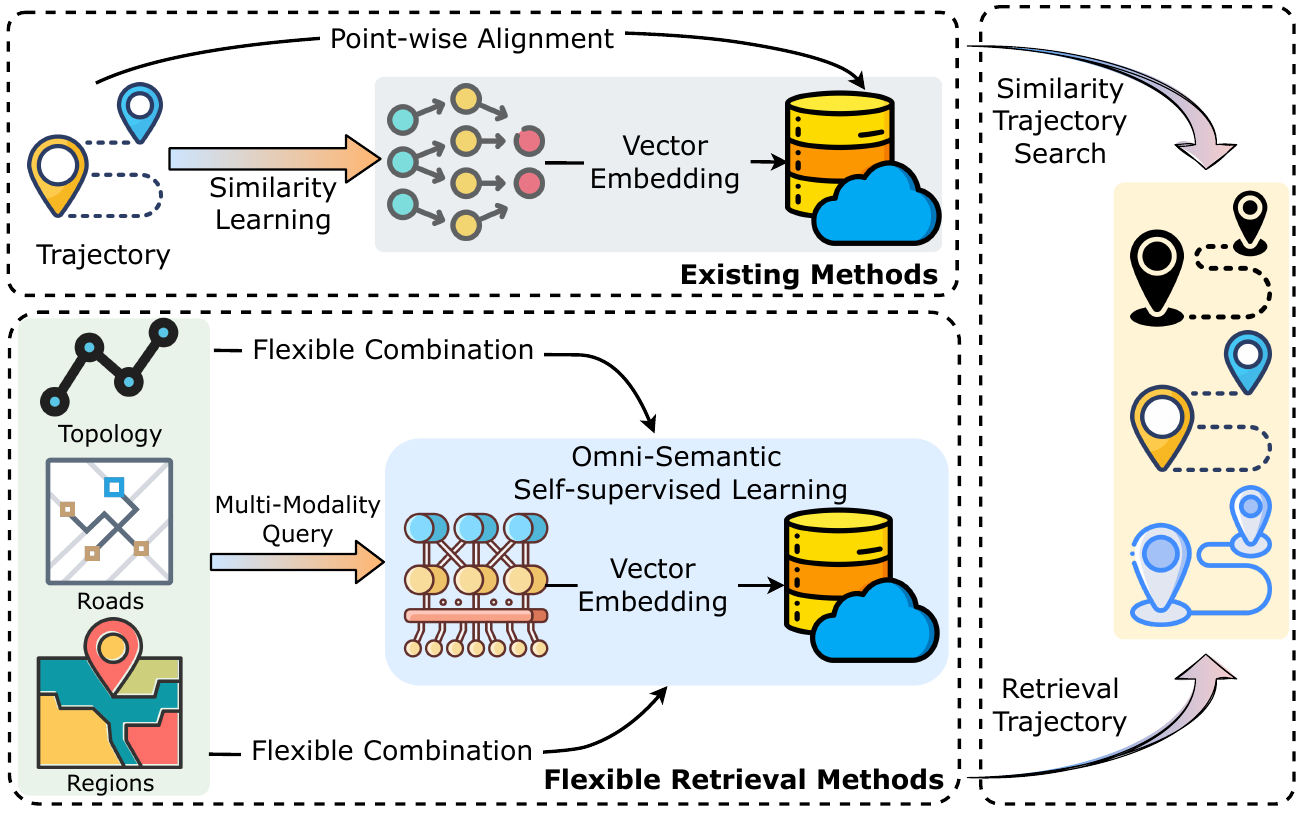}
    \vspace{-6mm}
    \Description[]{}
    \caption{Comparison of trajectory retrieval methods. Our method can retrieve trajectories according to their different modalities and provides flexible retrieval solutions.
    }
    \label{fig:intro}
    \vspace{-5mm}
\end{figure}

Existing approaches to trajectory retrieval focusing on trajectory similarity measures that can be broadly categorized into heuristic methods and deep learning-based methods, face critical limitations in handling \textit{complex queries}.
Heuristic methods, such as Dynamic Time Warping (DTW) \cite{muller2007dynamic} and Hausdorff Distance \cite{alt2009computational}, focus on computing the similarity between full trajectories based on point-wise alignment.
While effective for whole-trajectory matching, they suffer from prohibitive computational costs (e.g., $\mathcal{O}(n^2)$) time complexity for DTW), especially when applied to large-scale datasets and are limited to supporting only full-trajectory similarity queries.
Deep learning-based methods---including t2vec \cite{li2018deep} and TrajCL \cite{chang2023contrastive}---address efficiency issues by encoding trajectories into compact embeddings ($\mathcal{O}(n)$ time complexity when retrieving). 
However, these methods rely on holistic similarity computation and still do not address the support of feature-specific conditions like filtering trajectories that traverse designated road segments.
Recent multimodal trajectory studies \cite{lin2024trajfm,lin2023pre} have attempted to integrate ancillary data (e.g., road networks or POIs), but have primarily targeted specific classification or prediction tasks rather than cross-modal retrieval. 
For example, models inspired by contrastive learning \cite{zhou2024grid,xu2024mm} align trajectories with other descriptions, but lack a mechanism for querying trajectories based on spatial semantic combinations.
Overall, existing methods suffer from inefficiency in handling large-scale data, tunnel vision on trajectory similarity tasks, and unavailability to support conditional (or modality-based) queries. 
These restrictions hinder their applicability to real-world scenarios, where queries are inherently multi-faceted and demand both precision and efficiency at city scales.

To address these issues, we propose to integrate diverse trajectory modalities, such as trajectory, topology, road segments, and regions, into a unified framework, enabling flexible and modality-supported queries.
As shown in Figure \ref{fig:intro}, existing methods are limited to matching trajectories as a single modality for trajectory point-wise alignment or using deep learning models for similarity learning. 
Meanwhile, a multi-modality trajectory query needs to construct full semantic self-supervised models for multiple modalities to realize flexible retrieval based on user-specified conditions.
However, two fundamental challenges emerge in developing such a multi-modality retrieval model:
(1) \textbf{Semantic Disentanglement}. How to represent trajectories in a modality-specific embedding space (e.g., road segments vs. regions) while maintaining their interrelationships?
(2) \textbf{Flexibility and Efficiency}. How can multiple modalities during retrieval dynamically combine without sacrificing query efficiency while ensuring flexibility in supporting a wide range of complex queries?
For example, DTW-based methods may retrieve trajectories matching a road segment but ignoring regional constraints, while deep learning models fail to isolate modality-specific features, yielding irrelevant results.

\begin{table}[t]
\caption{Comparison of OmniTraj with representative trajectory retrieval/similarity  methods.}
\vspace{-3mm}
\centering
\small
\begin{tabular}{lcccc}
\toprule
Methods & DTW    & t2vec  & TrajCL  &  OmniTraj \\
\midrule
Retrieval Complexity  & $\mathcal{O}(n^2)$  & $\mathcal{O}(n)$  & $\mathcal{O}(n)$  & $\mathcal{O}(n)$ \\
Scalability  &   \xmark  & \cmark  & \cmark  &  \cmark \\
Conditional Query  & \xmark  & \xmark  & \xmark  &  \cmark \\
Multi-Modality Support & \xmark  & \xmark & \xmark  &  \cmark \\
Application Support &   \xmark   & \xmark  & \xmark  &  \cmark \\
\bottomrule       
\end{tabular}
\vspace{-4mm}
\label{tab:methods_comp}
\end{table}

To bridge existing research gaps and address the challenges, we propose \textbf{\model}, a generalized and flexible omni-semantic trajectory retrieval framework designed to integrate diverse trajectory aspects (e.g., topology, road segments, and regions) into a unified system.
Specifically,  the primary innovation of the \model is its ability to handle both spatial and semantic complementary constraints and provide a richer understanding of trajectory data. 
Firstly, OmniTraj encodes the unique features of each trajectory modality individually, ensuring that the precise description of each modality is preserved during the retrieval process and supports a wide range of applications. The framework then employs a modular design while preserving their interrelationships without sacrificing scalability or query efficiency. 
By decoupling the encoding process of each modality and dynamically combining them during retrieval, \model enables flexible and efficient queries.
In addition, the encoders for each modality can be used individually to support a wide range of downstream tasks.
As highlighted in Table \ref{tab:methods_comp}, OmniTraj outperforms traditional methods by flexible and scalable design, and supporting condition-based queries while maintaining high efficiency on large-scale datasets.
As a result, OmniTraj addresses the limitations of traditional trajectory similarity measures and provides a powerful solution that combines accuracy, efficiency, and versatility in trajectory retrieval.
By enabling the handling of complex queries based on a combination of trajectory, spatial, and semantic information, \model sets a new perspective for the trajectory retrieval task.

In summary, the contributions of our research are as follows:
\begin{itemize}[leftmargin=*]
    \item We formalize the novel task of multimodal trajectory retrieval, enabling condition-based queries that span diverse trajectory semantics. This approach overcomes the limitations of current trajectory similarity measures and provides a fresh insight into how trajectory retrieval tasks can be approached.

    \item We present OmniTraj, the first multimodal trajectory retrieval framework designed to process multiple trajectory modalities. This framework offers a generalized solution to trajectory retrieval by incorporating modality-specific encoders, enabling flexible, scalable, and efficient querying across a wide range of trajectory components.

    \item Through extensive experiments, we demonstrate the effectiveness of OmniTraj, showcasing its ability to accurately retrieve trajectories under diverse query conditions.
    In addition, the framework is scalable, supports a variety of downstream tasks, and demonstrates strong generalization capabilities.
\end{itemize}

\section{Preliminary}\label{sec:pre}
We formally define the multimodal trajectory retrieval task by introducing four key trajectory semantics, each providing distinct perspectives: trajectory, topology, road segments, and regions.

\noindent \textbf{Definition 1} \textbf{(Trajectory)}.
A trajectory $\boldsymbol{T}$ is a temporally ordered sequence of spatio-temporal points, defined as: $\boldsymbol{T} = \{p_1, p_2,~ \ldots,~ p_n \}$, where $p_i = (x_i,~y_i)$ represents the spatial coordinates (e.g., latitude and longitude) of the object (e.g., vehicle or pedestrian).
In real-world application, a trajectory can be abstracted into multiple granular representations---including \emph{topology}, \emph{road segments}, and \emph{regions}---to capture diverse semantic perspectives.

\noindent \textbf{Definition 2} \textbf{(Topology)}.
The topology $\boldsymbol{T}^{(\text{top})}$ of a trajectory $\boldsymbol{T}$ refers to the subsequence of critical points along the trajectory.
Formally,  $\boldsymbol{T}^{(\text{top})} = \{tp_1, tp_2, \ldots, tp_k\}, ~(k \leq n)$, where each  $tp_i \in \{p_1,~p_2, \ldots, p_n \}$ indicates a significant change in movement (e.g., abrupt turns, marked velocity variations, or crossings at predefined landmarks). 
This semantic emphasizes the spatial geometry of the trajectory while omitting temporal details.

\noindent \textbf{Definition 3} \textbf{(Road Segments)}.
For a trajectory that has been map-matched to a road network, its road segment representation is given by: $\boldsymbol{T}^{(\text{road})} = \{r_1,\, r_2, \ldots, r_j \}$, where each  $r_i$  is a unique identifier corresponding to a specific road segment (i.e., a continuous stretch of road between two intersections). 
This semantic captures the sequence of roads traversed by the trajectory.

\noindent \textbf{Definition 4} \textbf{(Regions)}.  
A region is a spatially bounded area with homogeneous properties, such as administrative boundaries, clusters of points of interest (POIs), or urban districts. 
The region semantic of a trajectory is defined as:
$\boldsymbol{T}^{(\text{reg})} = \{\mathcal{R}_1, \mathcal{R}_2, \ldots\}$,
where the condition $\boldsymbol{T} \cap \mathcal{R}_i \neq \varnothing$ indicates that at least one point of $\boldsymbol{T}$ lies within the region $\mathcal{R}_i$.

\noindent \textbf{Problem Statement} \textbf{(Trajectory Retrieval)}. 
Let $\mathcal{D} = \{\boldsymbol{T}_1, \ldots, \boldsymbol{T}_N\}$ be a database of trajectories. Trajectory retrieval aims to efficiently query and retrieve relevant trajectories from $\mathcal{D}$ based on one or more feature modalities (e.g., topology, road segments, regions, or their combination). 
Formally, let $T_q$ denote a query trajectory or query condition, which can be expressed using multiple modalities or combinations thereof. The retrieval function is defined as:
\begin{align}
\mathcal{F}(\boldsymbol{T}_q) = \arg \max_{\boldsymbol{T}_i \in \mathcal{D}} \text{similarity}(\boldsymbol{T}_q, \boldsymbol{T}_i),
\end{align}
where $\text{similarity}(\cdot~, \cdot)$ measures the degree of alignment or resemblance between the query $\boldsymbol{T}_q$ and a candidate trajectory $\boldsymbol{T}_i$ from the database. This similarity metric is designed to be highly versatile, potentially leveraging the information encoded in multiple modalities to provide a comprehensive comparison.
\emph{In practice, a robust trajectory retrieval framework should achieve high retrieval accuracy, offer flexibility for diverse query modalities, and efficiently manage large-scale databases with acceptable times.}

\section{Methodology}\label{sec:method}
\subsection{Overview of the OmniTraj Framework}
\begin{figure*}[!t]
    \includegraphics[width=0.98\linewidth]{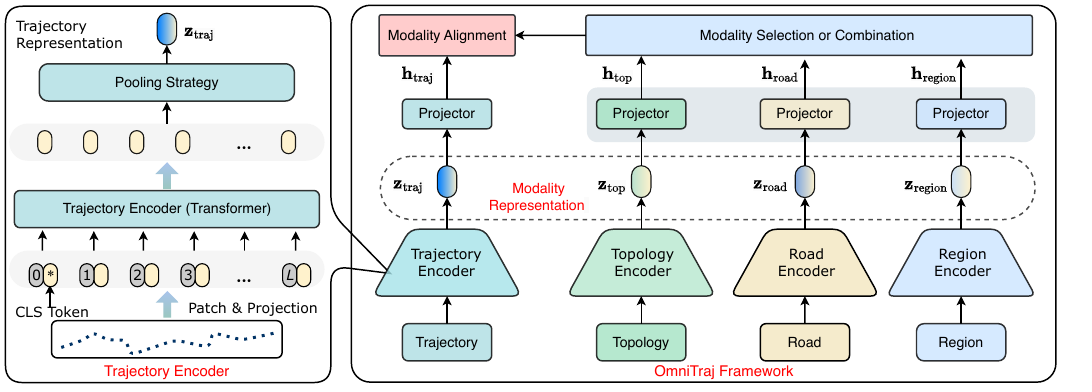}
    \Description[]{}
    \vspace{-3mm}
    \caption{Overview of the proposed OmniTraj framework, which contains four individual modal encoders. Each encoder captures a diverse semantic representation of a trajectory and projects it into a shared space.}
    \vspace{-3mm}
    \label{fig:overview}
\end{figure*}

To overcome the limitations of existing trajectory similarity computation methods and supporting multi-modality retrieval tasks, we propose OmniTraj, a novel omni-semantic trajectory retrieval framework that integrates four complementary modalities---raw trajectories, topology, road segments, and regions---into a unified system for flexible and efficient querying. 
As depicted in Figure~\ref{fig:overview}, OmniTraj is built upon four dedicated encoders, each meticulously designed to capture a distinct semantic aspect of trajectory data:
The \emph{Trajectory Encoder} processes the raw spatio-temporal sequence of each trajectory to generate latent representations that encapsulate movement dynamics and temporal evolution.
The \emph{Topology Encoder} identifies and embeds critical spatial landmarks (e.g., abrupt turns, speed changes) to characterize the geometric structure.
The \emph{Road Encoder} maps trajectories onto road network segments and learns road-aware embeddings through neural network architectures, effectively capturing connectivity and routing information.
The \emph{Region Encoder} detects and encodes the spatial-semantic properties of regions traversed by trajectories, thereby enriching the overall representation with contextual insights.
This modular design provides three principal advantages:
\begin{itemize}[leftmargin=*]
    \item \textbf{Flexible Query Support}: 
    OmniTraj empowers users to formulate queries using any combination of modalities. For instance, one may retrieve trajectories that pass through a specific road segment (e.g., Road 1) and intersect a particular region (e.g., Region B), enabling precise adaptation to diverse real-world scenarios.

    \item \textbf{Rich MultiModal Representations}: 
    Each encoder generates modality-specific embeddings that preserve the unique semantic characteristics of its respective modality. A contrastive learning strategy is then employed to align these embeddings within a shared latent space, ensuring cross-modal compatibility.
    
    \item \textbf{Scalability and Efficiency}: 
    The decoupled encoder architecture allows parallel processing of modalities, making the framework highly scalable and capable of efficiently handling large-scale trajectory databases.
    Additionally, when a modality-based query is required, it can be directly and smoothly integrated into existing frameworks without re-engineering.

\end{itemize}

\subsection{Modality-Specific Encoders}

In this section, we detail the construction of each encoder and describe how our design implements an efficient and flexible multimodal trajectory retrieval mechanism.
More details about the structure of these encoders can be found in \textbf{Appendix \ref{app:structure}} 

\subsubsection{\textbf{Trajectory Encoder}.}
Learning robust representations from raw spatio-temporal trajectories presents two key challenges: (1) capturing both local motion patterns and long-range spatial dependencies, and (2) accommodating variable sequence lengths while minimizing computational complexity for efficiency.
To tackle these challenges, we adopt a transformer-based architecture inspired by the Vision Transformer \cite{dosovitskiy2020image} to model the dynamic patterns.

Given a trajectory $\boldsymbol{T}$, we first normalize its coordinates to a local reference frame and up/down-sample to a fixed length $L$ using cubic spline interpolation, ensuring consistency across different scales or coordinate systems.
Then the resampled trajectory $\hat{\boldsymbol{T}}$ is divided into no-overlapping patches, where the number of patches is $N_p = \lfloor L/P \rfloor$ and $P$ is the patch size.
Each patch, denoted as $\boldsymbol{z}_p$, is computed by applying a learnable projection matrix $\boldsymbol{W}_p$ that maps the flattened patch coordinates into a $d$-dimensional latent space:
\begin{align}
\boldsymbol{z}_p = \boldsymbol{W}_p \cdot \hat{\boldsymbol{T}} \in \mathbb{R}^{N_p \times d},
\end{align}
To maintain the sequential order of the trajectory, we introduce learnable positional embeddings $\boldsymbol{E}_{\text{pos}} \in \mathbb{R}^{(N_p + 1) \times d}$ which are added to each patch embedding.
Additionally, a special [CLS] token $\boldsymbol{z}_{\text{cls}}$ is prepended to the sequence to aggregate global trajectory features. 
The full sequence $\boldsymbol{z} = [\boldsymbol{z}_{\text{cls}}; \boldsymbol{z}_p] + \boldsymbol{E}_{\text{pos}}$ is then processed by $N$ stacked transformer blocks. 
Each block computes multi-head self-attention over the spatial neighbors of each patch, allowing the model to learn both local motion patterns and long-range dependencies.
Finally, the trajectory embedding is derived from the [CLS] token or by pooling all patch embeddings. The trajectory representation is thus computed as:
\begin{align}
   \boldsymbol{z}_{\text{traj}} =\text{\textbf{TrajEncoder}}(T) \in \mathbb{R}^{ d}.
\end{align}
This encoder efficiently learns trajectory representations by capturing local patterns via patches, while also leveraging the self-attention mechanism to model long-range dependencies.
By using patches and projection, we reduce the computational complexity from the traditional $\mathcal{O}(L^2)$ to $\mathcal{O}((L/P)^2)$, significantly improving the efficiency of trajectory representation computation.
The resulting fixed-length trajectory embedding is well-suited for subsequent retrieval tasks, whether based on similarity or condition-based queries, offering both accuracy and computational efficiency solutions in large-scale settings.

\subsubsection{\textbf{Topology Encoder}.}
The main challenge in building a topology encoder is to capture the overall geometric semantics of the trajectory with only a limited number of key points.
This means that the encoder must efficiently process each point and derive a global spatial understanding from this sparse representation. 
To achieve this, given a topological sequence $\boldsymbol{T}^{(\text{top})}$ with $k$ points, the topology encoder first embeds each point  $tp_i \in \mathbb{R}^2$ into a high-dimensional vector $\{ \boldsymbol{e}_1, \boldsymbol{e}_2, \ldots, \boldsymbol{e}_k \}$ using a learned embedding layer.
To preserve the sequential order and capture the inherent spatial structure of these points, we incorporate Rotary Positional Embedding (RoPE) \cite{su2024roformer}. RoPE injects relative positional information by applying a position-dependent rotation to each embedding, thereby allowing the model to naturally learn the spatial dependencies between points.
Specifically, for each embedding $\boldsymbol{e}_i$, the rotation angle is $\theta_i={i}/{10000^{2k/d}}$, and the embedding is split into two halves,  $\boldsymbol{e}_i^{(1)}$ and $\boldsymbol{e}_i^{(2)}$.
The RoPE mechanism then rotates these splits as follows:
\begin{align}\label{eq:rope}
    \text{RoPE}(\boldsymbol{e}_i) = \begin{pmatrix}
    \cos \theta_i & -\sin \theta_i  \\
    \sin \theta_i & \cos \theta_i 
    \end{pmatrix} \begin{pmatrix}
    \boldsymbol{e}_i^{(1)} \\
    \boldsymbol{e}_i^{(2)}
    \end{pmatrix}.
\end{align}
After positional encoding, the sequence of RoPE-enhanced embeddings is processed by  $N$ transformer layers, which leverage self-attention mechanisms to model the interactions among sparse topological points. 
A special [CLS] token is prepended to the sequence to aggregate global geometric features. The final topological embedding is then obtained as:
\begin{align}
    \boldsymbol{z}_{\text{top}} =\text{\textbf{TopologyEncoder}}(\boldsymbol{T}^{(\text{top})}) \in \mathbb{R}^{ d}.
\end{align}
By leveraging RoPE, our topology encoder efficiently models sparse geometric structures and preserves critical spatial dependencies without the need for dense, point-wise embeddings.

\subsubsection{\textbf{Road Encoder}.}
The road encoder is designed to model the semantic and structural relationships between road segments in a map-matched trajectory, enabling flexible query conditions that go beyond exact matching.
Considering that our approach is not limited to precise road segment sequences, it should also support queries such as retrieving trajectories that pass through or cover a certain road segment sequence.
To facilitate this, we incorporate a series of augmentation techniques during training, which help the model learn robust and invariant representations. 
These augmentations include:
\begin{itemize}[leftmargin=*]
    \item Reversing: Flip the order of road segments to simulate bidirectional traversal.
    \item Discarding/Truncating: Randomly removing or retaining only part of the segments to mimic incomplete queries.
    \item Shuffling: Perturbing the order of segments within local windows to encourage the learning of invariant features.
    \item Random Replacement: Substituting some portions of segments with others to enhance generalization.
\end{itemize}
Considering that each road segment has a unique identifier, we first project each segment $r_i$ into a $d$-dimensional space using a learnable lookup table:
\begin{align}
\boldsymbol{z}_i = \boldsymbol{W}_\text{road}[r_i] \in \mathbb{R}^{d},
\end{align}
where $ \boldsymbol{W}_\text{road} \in \mathbb{R}^{r \times d}$ is the road embedding matrix and $|r|$ denotes the total number of unique road segments.
Next, we feed it into the $N$ transformer block enhanced with RoPE for processing to obtain the final road segment embedding:
\begin{align}
    \boldsymbol{z}_{\text{road}} =\text{\textbf{RoadEncoder}}(\boldsymbol{T}^{(\text{road})}) \in \mathbb{R}^{d}.
\end{align}
By integrating these augmentation techniques and advanced encoding strategies, the road encoder not only captures detailed structural information from the road network but also learns flexible semantic representations. 
This robustness enables the system to support a variety of condition-based queries and enhances its generalization to real-world retrieval scenarios, where queries might be incomplete, unordered, or require a degree of invariance.

\subsubsection{\textbf{Region Encoder}.}
Our motivation for constructing the region encoder is to model the spatial relationships and semantic features of the regions through which the trajectories pass, thereby providing robust and flexible representations of the paths of the trajectories in different spatial regions.
Similar to the road encoder, the region encoder supports condition-based queries by learning invariant and generalizable representations from region identifiers.
Specifically, each region identifier $\mathcal{R}_i$ is first projected into a high-dimensional embedding  $\boldsymbol{e}_i \in \mathbb{R}^d$ via a learnable lookup table, yielding a sequence of region embeddings $\{ \boldsymbol{e}_1, \boldsymbol{e}_2, \ldots, \boldsymbol{e}_m \}$. 
Unlike the road encoder, which emphasizes relative positional information, the region encoder processes this sequence of embeddings using a series of transformer blocks without explicitly incorporating positional information. 
The self-attention mechanism in the transformer naturally captures both local and global relationships among the regions, enabling the model to learn complex spatial dependencies.
To further enhance generalization and robustness, especially under conditions where region data may be incomplete or unordered, we apply augmentation techniques such as random shuffling and random removal of regions during training. 
These augmentations force the encoder to focus on the invariant semantic properties of the regions rather than their precise order.
Finally, the aggregated region trajectory embedding, denoted as:
\begin{align}
    \boldsymbol{z}_{\text{region}} =\text{\textbf{RegionEncoder}}(\boldsymbol{T}^{(\text{region})}) \in \mathbb{R}^{ d},
\end{align}
which can be obtained either by extracting the [CLS] token or by applying pooling over the sequence of region embeddings. 
This flexible aggregation strategy supports efficient querying based on region-specific conditions and ensures that the final representation effectively captures the semantic context of the trajectory.

\subsection{Omni-Semantic Supervision}
After obtaining the modality-specific embeddings $\boldsymbol{z}_{\text{mod}} \in \mathbb{R}^d$ and $ \boldsymbol{z}_{\text{mod}} \in \{\boldsymbol{z}_{\text{traj}}, ~\boldsymbol{z}_{\text{top}},~\boldsymbol{z}_{\text{road}},~\boldsymbol{z}_{\text{region}}\}$  from the trajectory, topology, road, and region encoders respectively, the next critical step in OmniTraj is to align these representations to supports flexible and efficient querying.
To achieve this, we designed a projection head employing a two-layer linear neural network for each modal representation.
This transformation is expressed as: $\boldsymbol{h}_{\text{mod}} = \boldsymbol{W}_{\text{mod}} \cdot \boldsymbol{z}_{\text{mod}}$, where $\boldsymbol{W}_{\text{mod}} \in \mathbb{R} ^{h \times d}$ is learnable projection matrix. 
This transformation ensures that the resulting embeddings  $\{ \boldsymbol{h}_{\text{traj}},~\boldsymbol{h}_{\text{top}},$  $ \boldsymbol{h}_{\text{road}},~\boldsymbol{h}_{\text{region}} \} \in \mathbb{R}^{h}$ all reside in the same latent space, making them directly comparable and combinable for downstream tasks.
To enforce consistency and semantic alignment across these different modalities, we employ a contrastive learning approach using the InfoNCE loss \cite{oord2018representation}. 
Although we adopt the simplest variant of the InfoNCE loss, it plays a crucial role in our framework by encouraging embeddings from the same trajectory to be close, while pushing apart those from different trajectories or modalities. 
Specifically, for a given trajectory $\boldsymbol{T}_q$  and its corresponding modality-specific representations $\boldsymbol{h}_{\text{mod}}$, the objective is to maximize the cosine similarity between the query embedding and its positive sample (i.e., the same embedding of trajectory from another modality), while minimizing the similarity with negative samples drawn from other trajectories. The InfoNCE loss is defined as:
\begin{align}
    \mathcal{L}_{\text{cont}} = - \log \frac{\exp(\text{sim}(\boldsymbol{h}_q,~\boldsymbol{h}_p) /\tau)}{\sum_i \exp(\text{sim}(\boldsymbol{h}_q,~\boldsymbol{h}_i)/\tau)},
\end{align}
where $\boldsymbol{h}_q$  is the query embedding, $\boldsymbol{h}_p$  is the positive sample embedding (i.e., the same semantic embedding from another modality),  $\boldsymbol{h}_i$  represents the negative samples. 
The temperature parameter $\tau$ controls the concentration of the distribution and $\text{sim}(\cdot, \cdot)$  denotes the cosine similarity calculated by: $\text{sim}(\boldsymbol{a}, \boldsymbol{b} ) = \frac{\boldsymbol{a}^{\top}\cdot \boldsymbol{b}} {\parallel \boldsymbol{a}\parallel  \parallel \boldsymbol{b}\parallel}.$
The contrastive learning framework is applied in both directions: from trajectory to modality and vice versa. This bidirectional alignment loss function is formalized as:
\begin{align}
   \mathcal{L} = \mathcal{L}_{\text{traj}\rightarrow\text{modality}} +\mathcal{L}_{\text{modality}\rightarrow\text{traj}}.
\end{align}
Beyond aligning single modalities, our design also supports the fusion of multiple modality representations. In practice, we can simply concatenate the required embeddings and pass them through a linear projector to produce a fused representation. 
This fused embedding inherits semantic cues from each individual modality, offering even richer representations for condition-based queries. The flexibility of this design allows the model to query using a single modality, any combination thereof, or a fully fused representation, thereby catering to diverse retrieval scenarios.

\subsection{Discussion}
\noindent \textbf{Framework Design Analysis}.
Our design is driven by both theoretical insights and practical requirements. We adopt a modular structure combined with the vanilla InfoNCE loss \cite{oord2018representation, radford2021learning} to project modality-specific embeddings into a shared latent space. This modular design not only facilitates efficient alignment across diverse inputs but also enables seamless integration of new modalities without extensive re-engineering. While more sophisticated contrastive losses (e.g., those with hard negative mining) or cross-attention mechanisms might offer further alignment improvements, we prioritize simplicity and scalability. The vanilla InfoNCE loss provides a proven, robust foundation that minimizes hyperparameter tuning and preserves computational efficiency—both critical factors for deployment in large-scale, real-world scenarios. In essence, our design achieves a generalized, flexible, and scalable framework that is well-suited to city-scale datasets.

\noindent \textbf{Time Complexity Analysis}.
The computational cost of OmniTraj mainly stems from two components: processing modality-specific embeddings through transformer layers and computing similarity scores. 
The self-attention in the transformer layers incurs a time complexity of $\mathcal{O}(N^2 \times d)$, where $N$ denotes the number of tokens (e.g., patches, segments, or regions) and $d$ is the embedding dimension. 
For similarity computation, the cost is $\mathcal{O}(|\mathcal{D}| \times n)$, where $|\mathcal{D}|$ represents the number of candidate embeddings and n is the number of query samples. 
Importantly, our use of vector similarity-based retrieval allows precomputation and storage of all candidate embeddings, avoiding redundant calculations and ensuring scalability and efficiency. 
In stark contrast, non-learned methods like DTW or Hausdorff scale quadratically $\mathcal{O}(|\mathcal{D}|^2 \times L^2)$ with the number of trajectories and sequence length $L$, rendering OmniTraj orders of magnitude faster for city-scale applications.

\section{Experiments}\label{sec:exper}

In this section, we first describe the datasets, baseline methods, and evaluation metrics used in our experiments. 
We then comprehensively evaluate the \model framework on two real-world datasets to answer the following research questions:
\begin{itemize}[leftmargin=*]
    \item \textbf{RQ1}: Can \model effectively retrieve trajectories compared to state-of-the-art trajectory similarity baselines?
    \item \textbf{RQ2}: How does \model perform when retrieving trajectories under specific conditions?
    \item \textbf{RQ3}: How do different parameter settings and architectural choices affect \model's performance?
    \item \textbf{RQ4}: Does \model exhibit transferability, scalability, and semantic understanding in real-world scenarios?
\end{itemize}

\subsection{Experimental Setups}

\subsubsection{\textbf{Datasets}}
We conducted experiments on two large-scale trajectory datasets collected from Chengdu and Xi'an, both widely recognized for their extensive coverage, high data quality, and frequent use in trajectory analysis studies \cite{chang2023contrastive,xu2024mm}. 
For each dataset, we reserve 20,000 trajectories for testing and 50,000 for validation, while approximately 1.1 million trajectories are used for training. 
Detailed statistics and descriptions are provided in \textbf{Appendix \ref{app:dataset}}.

\subsubsection{\textbf{Implement Details}}
The \model implementation involves four dedicated modality encoders and hyperparameters. 
The implementation code is provided online\footnote{https://github.com/Yasoz/OmniTraj}.
See \textbf{Appendix \ref{app:structure}} for details.

\subsubsection{\textbf{Baselines}}
We evaluate \model on two distinct retrieval tasks---trajectory similarity retrieval and condition-based retrieval---by comparing it against two groups of baselines.
For trajectory similarity retrieval, we consider two types of baselines. First, we adopt four classical heuristic algorithms like DTW \cite{muller2007dynamic}, EDR \cite{chen2005robust}, Hausdorff \cite{alt2009computational}, and Fréchet \cite{alt1995computing}, which offer well-established distance or similarity measures between trajectories. 
Second, we incorporate four state-of-the-art self-supervised learning approaches like E2DTC \cite{fang20212}, t2vec \cite{li2018deep}, TrjSR \cite{cao2021accurate}, and TrajCL \cite{chang2023contrastive}, which learn compact embeddings for trajectory representation. 
During our evaluation, these methods were applied using topology modality to calculate the similarity with the target trajectory.
\emph{For condition-based retrieval, since no existing work directly addresses this task}, we adopt a set of simple multimodal approaches (e.g., basic spatial or semantic alignment techniques) as baselines. 
Further details regarding these baselines can be found in \textbf{Appendix \ref{app:baseline}}.

\subsubsection{\textbf{Evaluation Metrics}}
For trajectory similarity retrieval tasks, we use standard ranking metrics, including Mean Rank (\textbf{MR}), Mean Reciprocal Rank (\textbf{MRR}), and Hit Rate (\textbf{HR@k}) at various cutoff values,  to assess the effectiveness of the retrieval.
For the condition-based retrieval task, we employ the coverage metric (\textbf{CR}) to measure the accuracy of condition matching. 
Specifically, we report \textbf{CR@1} and \textbf{CR@5}, which represent 
the proportion of query conditions correctly matched in the top-1 and top-5 retrieved results, respectively. Additional details can refer to \textbf{Appendix \ref{app:metrics}}.

\subsection{Comparison to Trajectory Similarity Models}

\begin{table}[t]
\centering
\caption{Comparison of trajectory retrieval performance with similarity computation methods.}
\vspace{-3mm}
\resizebox{1\linewidth}{!}{
\begin{tabular}{l|lcccc}
\toprule
\textbf{Dataset} & \textbf{Method} & \textbf{MR}  & \textbf{MRR} & \textbf{HR@1}  & \textbf{HR@10} \\
\cmidrule(lr){1-6} 
\multirow{14}{*}{Chengdu} 
    & DTW & 23.051 & 0.522 & 0.402 & 0.771 \\ 
    & EDR & 16.15 & 0.279 & 0.137 & 0.587 \\ 
    & Hausdorff & 5.643 & 0.760 & 0.682 & 0.897 \\ 
    & Fréchet & 12.659 & 0.466 & 0.336 & 0.728 \\ 
    & E2DTC & 8.394 & 0.498 & 0.447 & 0.670 \\ 
    & t2vec & 11.474 & 0.513 & 0.354 & 0.638  \\ 
    & TrjSR & 6.472 & 0.568 & 0.536 & 0.795 \\ 
    & TrajCL & 1.463 & 0.846 & 0.791 & 0.974 \\ 
    & \model(reg) & 2.811 & 0.474 & 0.352 & 0.756 \\ 
    & \model(reg+top) & 1.441 & 0.825& 0.762 & 0.932 \\ 
    & \model(road) & 1.473 & 0.839 & 0.760 & 0.967 \\ 
    & \model(road+top) & 1.394 & 0.845 & 0.785 & 0.944 \\ 
    & \model(reg+road+top) & 1.401 & 0.843 & 0.782 & 0.943 \\ 
    & \model & \textbf{1.280} & \textbf{0.909} & \textbf{0.857} & \textbf{0.989} \\ 
\cmidrule(lr){1-6} 
\multirow{14}{*}{Xi'an} 
    & DTW & 35.651 & 0.391 & 0.273 & 0.608 \\
    & EDR & 11.539 & 0.339 & 0.191 & 0.669 \\
    & Hausdorff & 5.364 & 0.774 & 0.696 & 0.927 \\
    & Fréchet & 12.740 & 0.419 & 0.283 & 0.693 \\
    & E2DTC & 9.592 & 0.453 & 0.406 & 0.658 \\ 
    & t2vec & 10.143 & 0.417 & 0.364 & 0.664 \\ 
    & TrjSR & 7.811 & 0.579 & 0.532 & 0.816 \\ 
    & TrajCL & 1.492 & 0.832 & 0.805 & 0.968 \\ 
    & \model(reg) & 2.058 & 0.695 & 0.659 & 0.927 \\ 
    & \model(reg+top) & 1.450 & 0.860 & 0.783 & 0.983 \\ 
    & \model(road) & 1.434 & 0.865 & 0.795 & 0.981 \\ 
    & \model(road+top) & 1.319 & 0.897 & 0.840 & 0.987\\ 
    & \model(reg+road+top) & 1.326 & 0.896 & 0.838 & 0.986 \\ 
    & \model & \textbf{1.304} & \textbf{0.903} & \textbf{0.847} & \textbf{0.990} \\ 
\bottomrule
\end{tabular}
\label{tab:sim_results}
}
\vspace{-5mm}
\end{table}

Table \ref{tab:sim_results} summarizes the performance of various trajectory retrieval or similarity computation methods on the Chengdu and Xi’an datasets. 
Overall, the results demonstrate that the \model consistently outperforms both classical heuristic algorithms and state-of-the-art learning-based approaches in the trajectory similarity retrieval task, particularly when using the topology modality. 
This superior performance is primarily attributable to \model’s innovative architecture, which effectively encodes and aligns the complementary semantics of trajectory and topology, thereby enabling a fine-grained understanding of spatial dynamics.
Among the learning-based methods, TrajCL shows competitive results; However, \model achieves significantly higher HR@1 values, indicating a markedly improved ability to retrieve the most relevant trajectory as the top result. 
This above result suggests that while contrastive learning methods like TrajCL are effective for embedding trajectories, the multi-modality alignment strategy employed by \model leads to more precise and reliable retrieval outcomes. 
Moreover, the incorporation of topology modalities further enhances matching accuracy.

We further evaluate individual modalities by comparing \model(reg) and \model(road) to their combinations with topology (\model(reg+top), \model(road+top), and all of them \model(reg+road+top)). 
The results indicate that although both road segment and region semantics provide valuable contextual information, they offer a relatively coarse-grained description of trajectories, e.g., many trajectories may pass through the same road segments or regions, limiting their ability to capture detailed spatial movements. 
When we gradually refine the granularity of retrieval or fuse more accurate modalities, the retrieval accuracy can be further improved.
Nevertheless, these modalities are still helpful for rapidly narrowing down the candidate set in a multi-stage retrieval process: coarse filtering using region or road information can be followed by fine-grained topology-based matching (we add this extra experiment in \textbf{Appendix \ref{app:exp_twostage}}). 
This two-stage approach can reduce computational overhead and also improve overall efficiency, especially when scaling to city-level datasets.
Overall, these findings validate our claims that the \model is a flexible, scalable, and robust framework for trajectory retrieval, demonstrating high accuracy and generalizability across different urban environments.

\begin{table}[t]
\centering
\caption{Comparison of condition-based retrieval performance with modality alignment methods.}
\vspace{-3mm}
\resizebox{1\linewidth}{!}{
\begin{tabular}{l|lcccc}
\toprule
\multirow{2}{*}[\multirowoffset]{Dataset} & \multirow{2}{*}[\multirowoffset]{Method} & \multicolumn{2}{c}{\textbf{Road}} & \multicolumn{2}{c}{\textbf{Region}} \\
\cmidrule(lr){3-4}\cmidrule(lr){5-6}
& & \textbf{CR@1}  & \textbf{CR@5} & \textbf{CR@1}  & \textbf{CR@5}  \\
\cmidrule(lr){1-6} 
\multirow{7}{*}{Chengdu} 
& Embedding &  0.103   & 0.074 & 0.225  & 0.185   \\
& Linear &  0.148 &	0.130  &  0.283 &	0.261  \\
& CLIP (LSTM)  &  0.906   & 0.469  & 0.837   & 0.653  \\
& CLIP (no-aug) &  0.941   & 0.594   & 0.970   & 0.848  \\
& CLIP &  0.969   & 0.645 & 0.975   & 0.906 \\
& OmniTraj (no-aug) &  0.981   & 0.570      & 0.956   & 0.853  \\
& OmniTraj &  0.989   & 0.670     & 0.989   & 0.915  \\
\cmidrule(lr){1-6} 
\multirow{7}{*}{Xi'an} 
& Embedding &  0.087  & 0.054  & 0.210    & 0.187  \\
& Linear & 0.118 & 0.064  & 0.260 & 0.202  \\
& CLIP (LSTM) &  0.891   & 0.518   & 0.925   & 0.737  \\
& CLIP (no-aug) &  0.946   & 0.570   & 0.968   & 0.803  \\
& CLIP &   0.935   & 0.647   & 0.978   & 0.867  \\
& OmniTraj (no-aug) &  0.945   & 0.537 & 0.973   & 0.823  \\
& OmniTraj &  0.987   & 0.676   & 0.994   & 0.893  \\
\bottomrule
\end{tabular}
}
\vspace{-5mm}
\label{tab:sim_results_condition}
\end{table}

\subsection{Condition-based Retrieval Performance}

Table \ref{tab:sim_results_condition} presents the performance of various condition-based retrieval methods on the Chengdu and Xi’an datasets, evaluated using the coverage metrics CR@1 and CR@5 for both the road and region modalities. The results clearly demonstrate that the \model not only excels in trajectory similarity retrieval but also significantly enhances the matching accuracy between query conditions and the retrieved trajectories. 
In particular, \model achieves the highest CR@1 scores across both datasets and modalities, outperforming baseline methods and indicating that a large proportion of query conditions are met in the top retrieval results.
Notably, the robust performance of \model is maintained regardless of the encoder architecture used, whether LSTM or BERT in the CLIP approach, highlighting the flexibility and adaptability of our design. Moreover, the augmented version of \model consistently outperforms its unaugmented counterpart, especially in terms of CR@5. 
This improvement suggests that data augmentation substantially broadens the coverage of relevant trajectories, making the retrieval process more effective at identifying a wide range of candidates. Such an emphasis on coverage is critical for condition-based retrieval tasks, where the objective is to retrieve as many pertinent samples as possible rather than achieving pinpoint precision.
Overall, these findings underscore that \model’s multimodal alignment strategy captures complex semantic relationships between queries and target trajectories and supports flexible, high-coverage retrieval across diverse conditions. 
We also provide an analysis of the coverage changes with the length of the sequences and the range of the retrieved in \textbf{Appendix \ref{app:exp_condition}}

\begin{figure}[t]
\centering
    \subfigure[Model Structure.]{
        \includegraphics[width=0.48\linewidth]{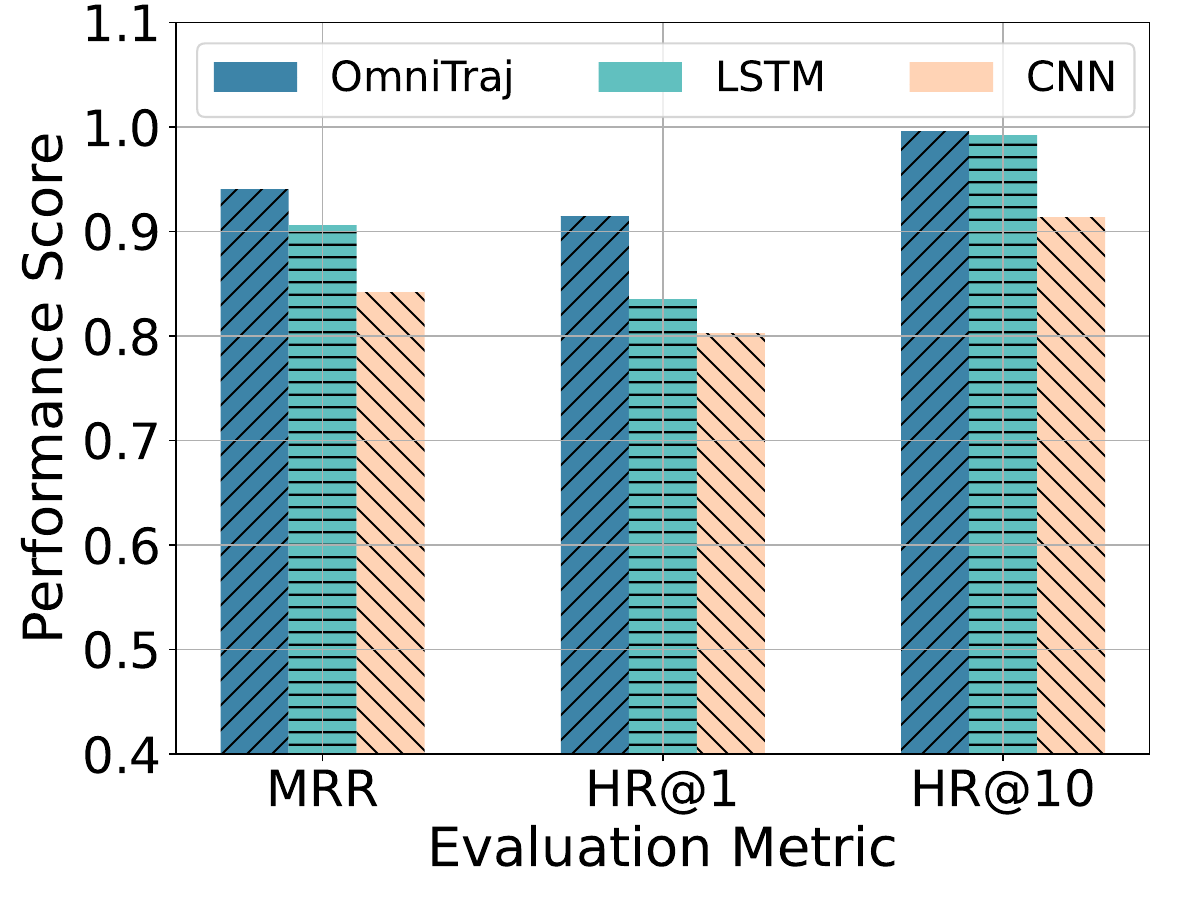}
        \label{fig:model_structure}
    }  \hspace{-0.05\linewidth} 
    \subfigure[Number of Encoders.]{
        \includegraphics[width=0.48\linewidth]{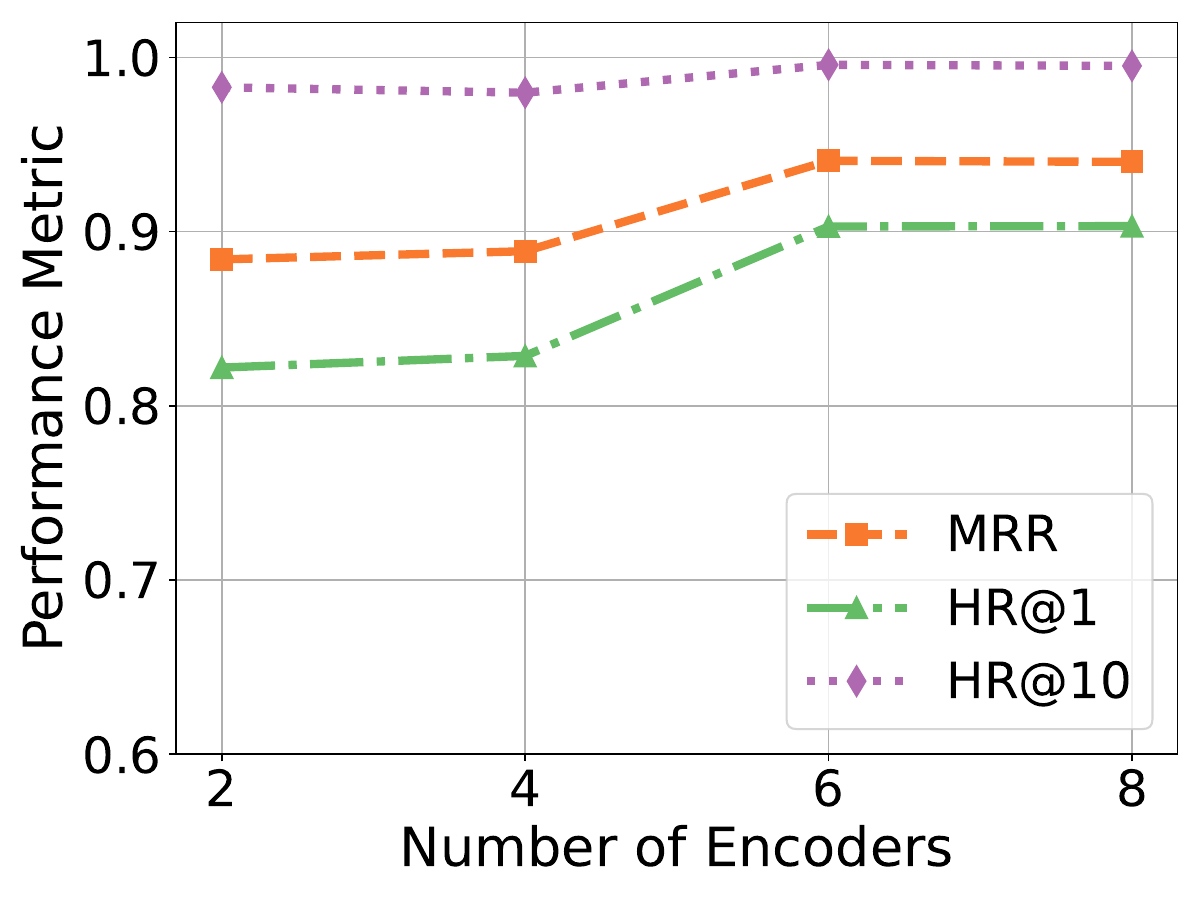}
         \label{fig:encoders}
    }
    \Description[]{}
    \vspace{-5mm}
   \caption{Model structure and parameters and setting.}
   \label{fig:RQ3}
    \vspace{-3mm}
\end{figure}

\subsection{Architecture Study and Parameters Setting}
Furthermore, we examine the influence of different architectural choices and the number of encoder blocks on \model’s performance. Figure \ref{fig:model_structure} compares the original OmniTraj architecture with alternative designs based on LSTM and CNN. The results clearly indicate that the original OmniTraj architecture delivers superior performance, significantly surpassing the LSTM and CNN variants. As we transition from OmniTraj to the other architectures, a gradual decline in retrieval accuracy is observed, underscoring the critical role of an effective multimodal encoding strategy. In particular, the combination of sequence modeling and attention mechanisms in OmniTraj enables a more robust alignment and better captures the underlying semantic relationships than traditional sequence-based models.
Figure \ref{fig:encoders} further investigate the effect of varying the number of encoder blocks. 
As the number of blocks increases from 2 to 6, there is a steady improvement in retrieval performance, suggesting that additional layers enhance the model’s capacity to capture and align complex multimodal representations. Notably, even with only two encoder blocks, the model achieves relatively good performance, demonstrating the inherent strength of our design. However, when the number of encoder blocks exceeds 6, the performance gains plateau, indicating that the model reaches a saturation point where further increases in depth yield diminishing returns. The above finding highlights the importance of balancing model complexity and computational efficiency, as increasing the number of encoder blocks beyond a certain threshold yields diminishing returns.

\subsection{Model Applicability Analysis}

In this section, we analyze the applicability of the \model by examining its transferability, scalability, and semantic understanding.

\begin{figure}[t]
\centering
    \subfigure[Transfer learning.]{
        \includegraphics[width=0.48\linewidth]{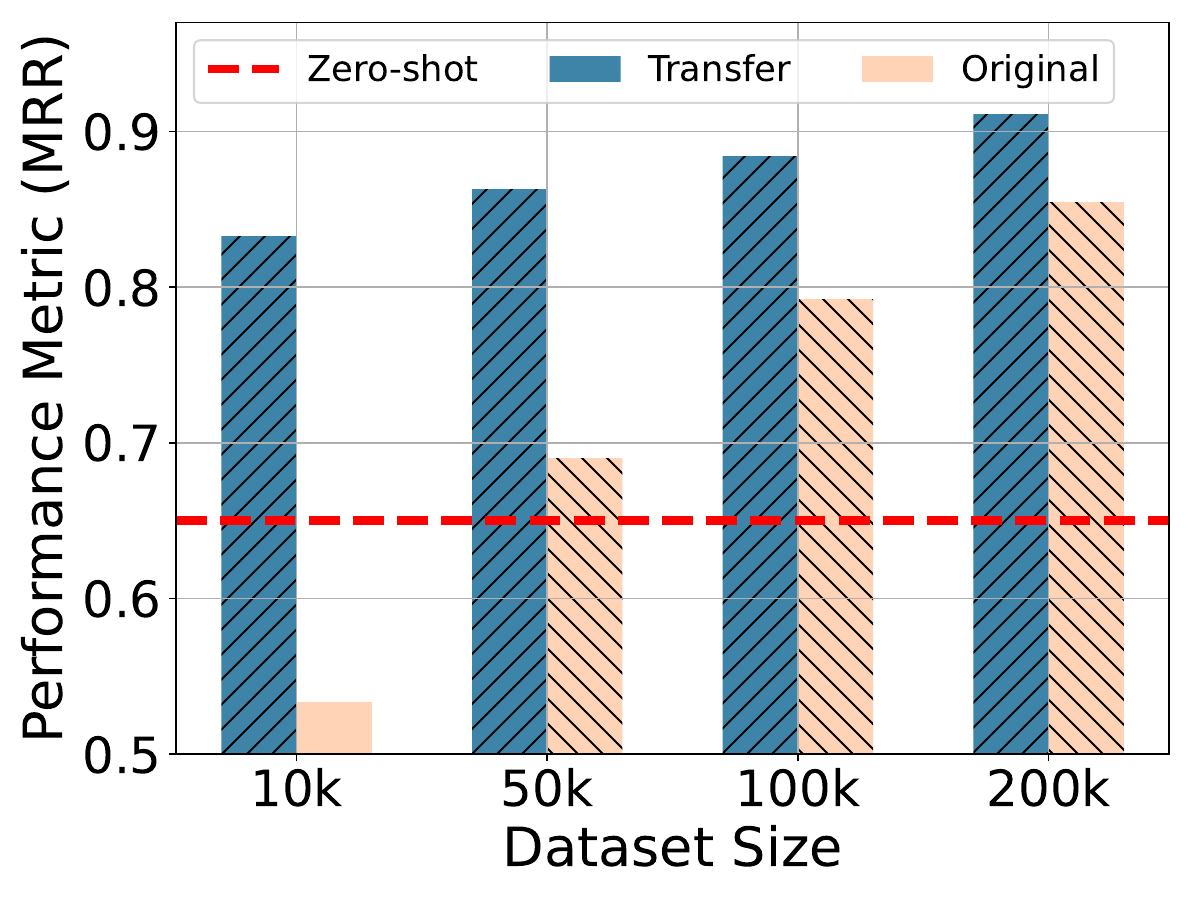}
        \label{fig:transfer}
    }  \hspace{-0.05\linewidth} 
    \subfigure[Data scale impact.]{
        \includegraphics[width=0.48\linewidth]{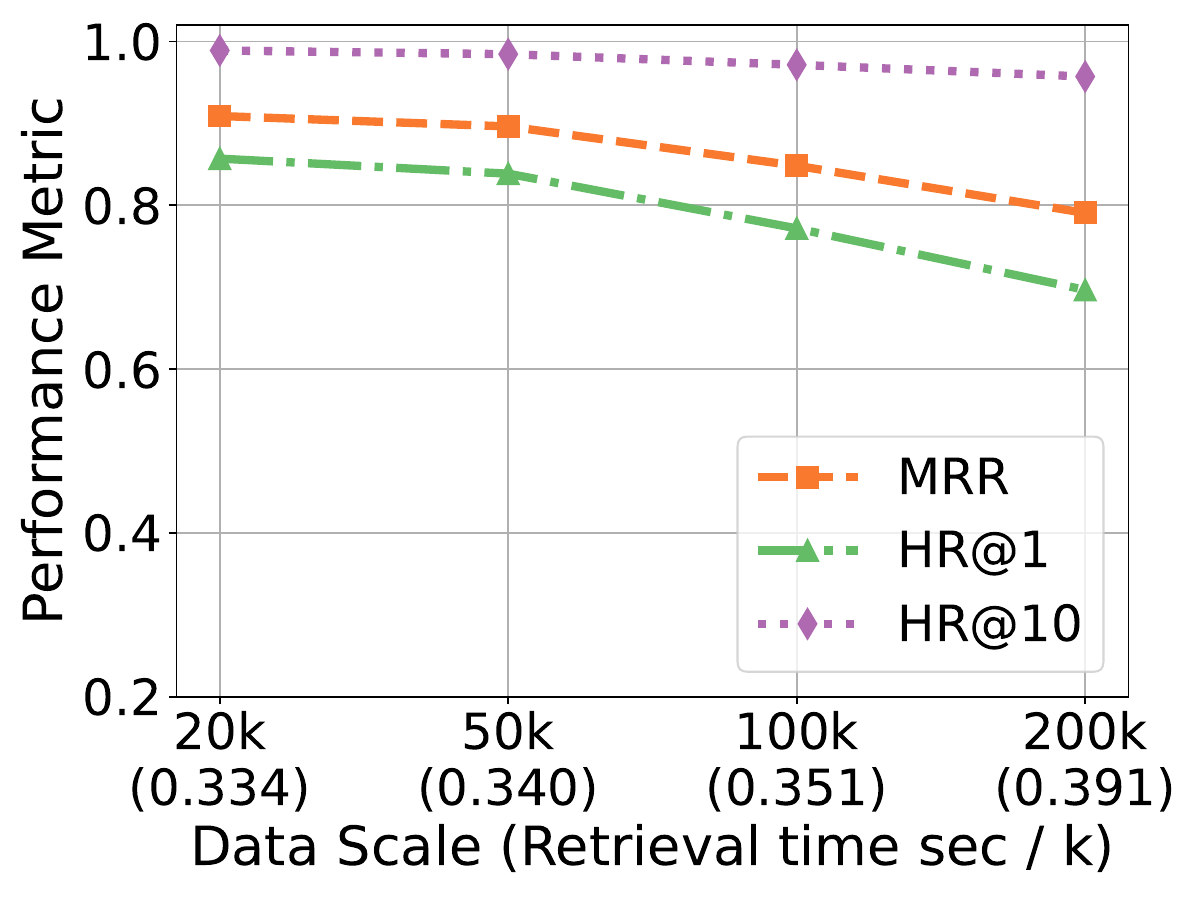}
        \label{fig:datascale}
    }
    \Description[]{}
    \vspace{-5mm}
   \caption{Model transferability and scalability analysis.}
    \vspace{-3mm}
   \label{fig:RQ4}
\end{figure}

\noindent \textbf{Transfer Learning}.
We first evaluate the zero-shot performance of the model and its ability to transfer between urban environments. 
Specifically,  we trained a model on data from Chengdu city and then directly applied it to a Xi'an city, examining robust zero-shot capability. 
To further assess transferability, we fine-tune the pre-trained model on the target city using varying amounts of training samples (10k, 50k, 100k, and 200k trajectories). 
As shown in Figure \ref{fig:transfer}, testing directly with pre-trained models in a new environment can also achieve comparable zero-shot performance to training with a 50k data size.
This is due to the fact that our model encodes topological modalities without adding any additional city-related information, allowing it to have better generalization capabilities.
The above statement is further verified by transfer learning. 
It clearly observes that even with limited data in the target domain, the performance of the model improves significantly  (``transfer'' for applying transfer learning with a pre-trained model and ``origin'' for direct training from scratch).
As the amount of data increases, the gap between the two models also reduces as the amount of data increases, validating the applicability of our \model to the new urban environment.

\noindent \textbf{Scalability}.
We check the scalability of the model by analyzing how the retrieval performance and processing time of the model change as the size of the dataset increases. 
We gradually increase the number of trajectories used for retrieval (20k, 50k, 100k, and 200k) and observe that although the overall retrieval performance decreases slightly with the increase in data size, the decrease is within acceptable limits.
At the same time, the processing time per 1,000 query samples gradually increases from 0.334 seconds at 20k to 0.391 seconds at 200k. 
This smooth trend reflects a reasonable trade-off between performance and efficiency when the model is scaled to handle larger datasets and can be effectively applied to large-scale data retrieval.

\begin{figure}[t]
\centering
    \subfigure[Topology.]{
        \includegraphics[width=0.31\linewidth]{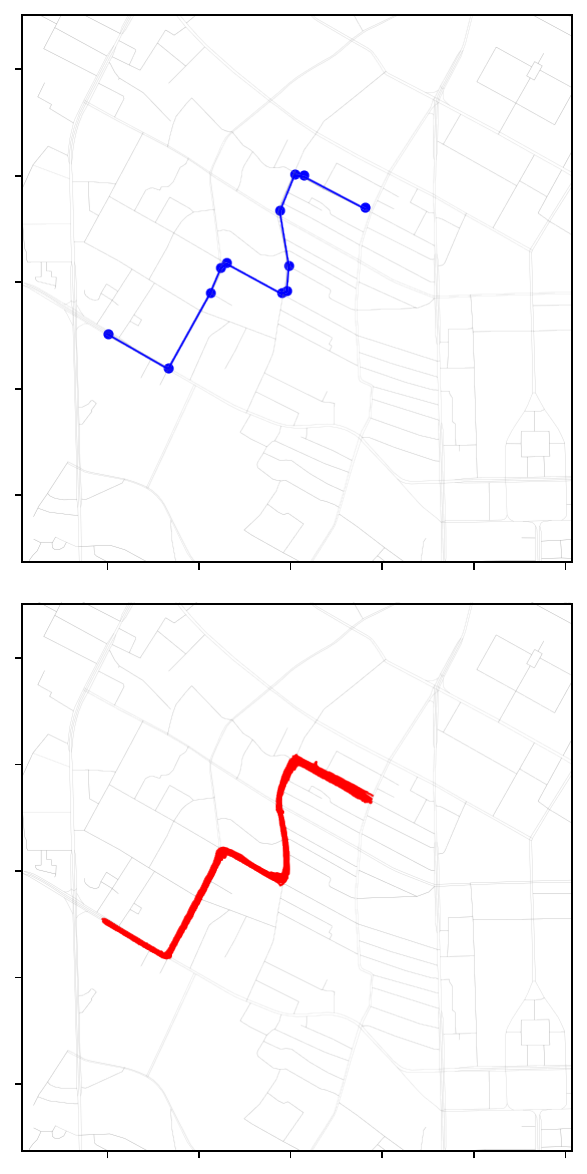}
    }  \hspace{-0.025\linewidth} 
    \subfigure[Road.]{
        \includegraphics[width=0.31\linewidth]{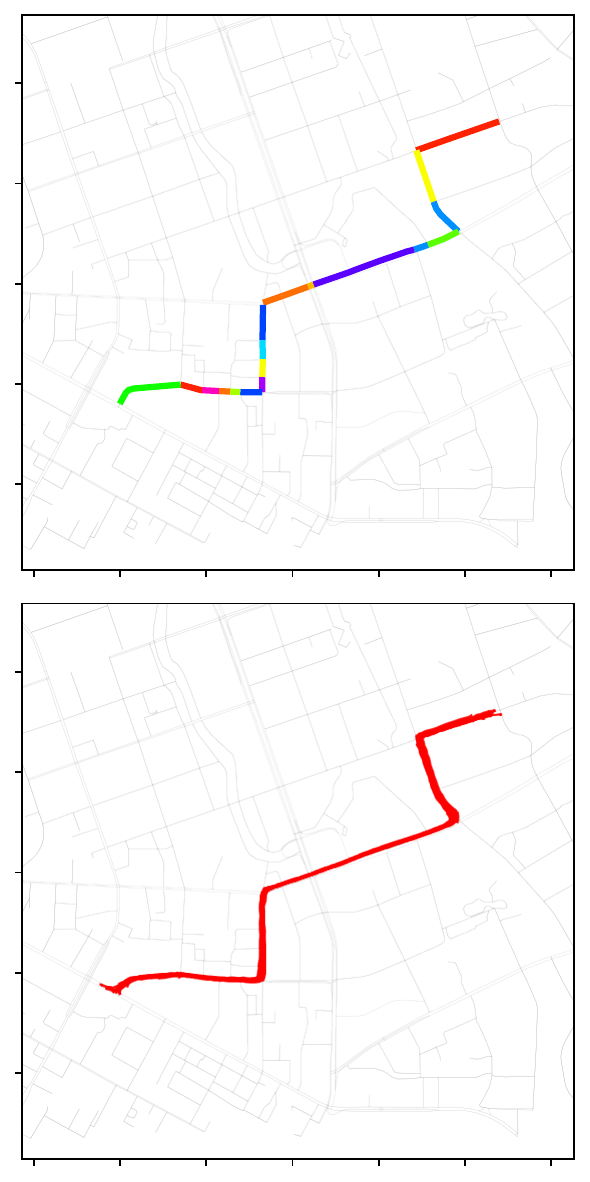}
    }
    \hspace{-0.025\linewidth}
    \subfigure[Region.]{
        \includegraphics[width=0.31\linewidth]{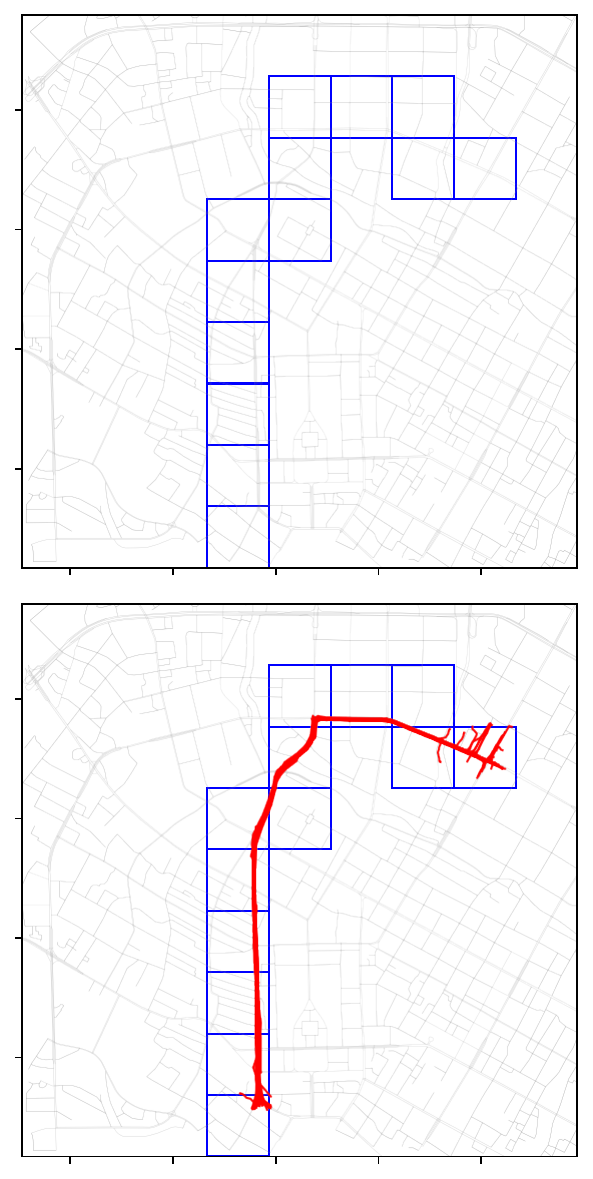}
    }
    \Description[]{}
    \vspace{-3mm}
   \caption{Use OmniTraj as a semantic representation to guide the model for trajectory generation.}
    \vspace{-3mm}
   
   \label{fig:exp_generation}
\end{figure}

\noindent \textbf{Semantic Understanding}.
To assess \model's semantic understanding and its utility for downstream applications, we conducted a pioneering trajectory generation task.
These encoders are then used to generate semantic embeddings that serve as conditional guidance inputs for the trajectory generation model \cite{zhu2024controltraj}.
As vividly illustrated in Figure \ref{fig:exp_generation}, the generated trajectories consistently and accurately match the specified conditions. 
This clear visualization unequivocally confirms that \model effectively captures and utilizes the intricate semantic information embedded within different modalities.
The capability underscores \model's profound understanding of the underlying spatial and semantic relationships within trajectories. 
onsequently, it demonstrates \model's robust potential to successfully support a wide range of downstream applications, including but not limited to targeted trajectory generation, intelligent route planning, and other spatially constrained tasks.

Overall, the above experiments validate that the \model is flexible and scalable and also exhibits strong transferability and semantic understanding, making it well-suited for real-world, large-scale trajectory retrieval and related applications.

\balance
\section{Related work}\label{sec:related}
\noindent \textbf{Trajectory Retrieval and Querying}.
Trajectory retrieval seeks to identify trajectories from a database that meet specific query criteria, such as similarity to a given trajectory or adherence to particular feature-based conditions \cite{zheng2019reference,tedjopurnomo2021similar}. 
This task is fundamental to spatio-temporal data analysis and underpins applications in intelligent transportation, urban analytics, and mobility recommendation \cite{dai2015personalized,de2022personalized}. 
Most existing research emphasizes trajectory similarity measures, where the goal is to match a given query trajectory with candidate trajectories in terms of spatio-temporal closeness \cite{chen2020parallel}.
Traditionally, trajectory similarity is computed using distance measures (e.g., Dynamic Time Warping \cite{muller2007dynamic}, Hausdorff \cite{alt2009computational}, or Fréchet \cite{alt1995computing}) or accelerated via indexing structures (e.g., grid-based partitions \cite{chang2023contrastive, cao2021accurate,li2018deep} or road network mappings \cite{yao2022trajgat,han2021graph}). 
However, these methods focus on matching entire trajectories and require processing all points, making them computationally expensive and ill-suited for flexible, feature-specific queries.
More recent deep learning approaches have leveraged trainable embeddings---using RNN-based encoders or attention mechanisms---to capture complex semantic and motion dynamics \cite{li2018deep,yao2019computing,zhou2023grlstm}, converting trajectories into compact embedding vectors that improve matching accuracy and efficiency. Nevertheless, these techniques largely address whole-trajectory similarity measures and offer limited support for partial or condition-based retrieval.
Condition-based retrieval, in contrast, aims to retrieve trajectories that satisfy specific constraints. 
While traditional indexing techniques can perform simple filtering, deep learning solutions have yet to fully integrate diverse trajectory features into a unified representation that supports precise condition-based queries \cite{zheng2015approximate,wang2022road,gao2020semantic}. 
This gap highlights the need for novel frameworks that reconcile the strengths of deep embeddings with the demands of multi-aspect trajectory retrieval.

\noindent \textbf{Multimodal Learning}.
Multimodal learning, which involves integrating and processing information from multiple data sources or modalities, has emerged as a pivotal area in machine learning, significantly advancing fields such as Computer Vision (CV) and Natural Language Processing (NLP) \cite{xu2023multimodal,yuan2025survey,deng2022multi}. 
Taking models such as CLIP as an example, a multimodal approach that combines image and textual descriptions enhances the robustness of visual representations and semantic understanding, while producing context-aware language models \cite{radford2021learning}.
In addition, combining text data with other modalities, such as video and image inputs, can align the semantic features of different modalities to achieve cross-modal retrieval and querying \cite{luo2022clip4clip}.
Extending these advancements to trajectory data, multimodal learning facilitates the comprehensive analysis of trajectories by incorporating diverse aspects such as road segments, street views, and regional POI information \cite{xu2024mm,lin2024unite}.
For example, the existing work incorporates multiple embeddings through a multi-task learning framework by combining trajectories with the rest of the corresponding modalities, which can effectively facilitate the trajectory-based representation learning task.
However, existing multimodal learning approaches primarily focus on enhancing target representations through the fusion of multiple modalities, thereby improving tasks like classification, prediction, and anomaly detection \cite{zhou2024grid,lin2024trajfm}. 
There is a noticeable gap in addressing multimodal retrieval, where the goal is to enable cross-modal or condition-based queries across different trajectory aspects. 
This limitation underscores the novelty of our work, which seeks to bridge this gap by developing a unified retrieval framework that leverages multimodal representations for flexible trajectory query.

\section{Conclusion}\label{sec:conclusion}

In this work, we introduce a new task in multimodal trajectory retrieval that enables condition-based queries across different trajectory semantics, including topology, road segments, and regions. 
To achieve this target, we propose OmniTraj, a unified omni-semantic trajectory retrieval framework designed to offer efficiency, flexibility, and comprehensive support for complex queries.
By leveraging semantic decoupling and a modular encoder design, OmniTraj ensures accurate and scalable retrieval while supporting applications that operate under various semantic constraints. 
This innovative approach addresses the limitations of traditional trajectory similarity measures, providing a fresh perspective on how trajectory retrieval tasks can be tackled.
Future work will extend OmniTraj to dynamic environments and real-time trajectory streams, further enhancing the generalization and flexibility of the proposed model.

% \section{Acknowledgment}
% This work is mainly supported by the National Natural Science Foundation of China (No. 62402414). This work is also supported by the Guangdong Basic and Applied Basic Research Foundation (No. 2025A1515011994), Guangzhou Municipal Science and Technology Project (No. 2023A03J0011), the Guangzhou Industrial Information and Intelligent Key Laboratory Project (No. 2024A03J0628), and a grant from State Key Laboratory of Resources and Environmental Information System, and Guangdong Provincial Key Lab of Integrated Communication, Sensing and Computation for Ubiquitous Internet of Things (No. 2023B1212010007).
% This research was partially supported by Research Impact Fund (No.R1015-23), Collaborative Research Fund (No.C1043-24GF), Huawei (Huawei Innovation Research Program, Huawei Fellowship), Tencent (CCF-Tencent Open Fund, Tencent Rhino-Bird Focused Research Program), Alibaba (CCF-Alimama Tech Kangaroo Fund No. 2024002), Ant Group (CCF-Ant Research Fund), and Kuaishou.

\clearpage
\bibliographystyle{ACM-Reference-Format}
\bibliography{ref}

\clearpage
\begin{appendix}
\begin{figure*}[!t]
    \includegraphics[width=0.98\linewidth]{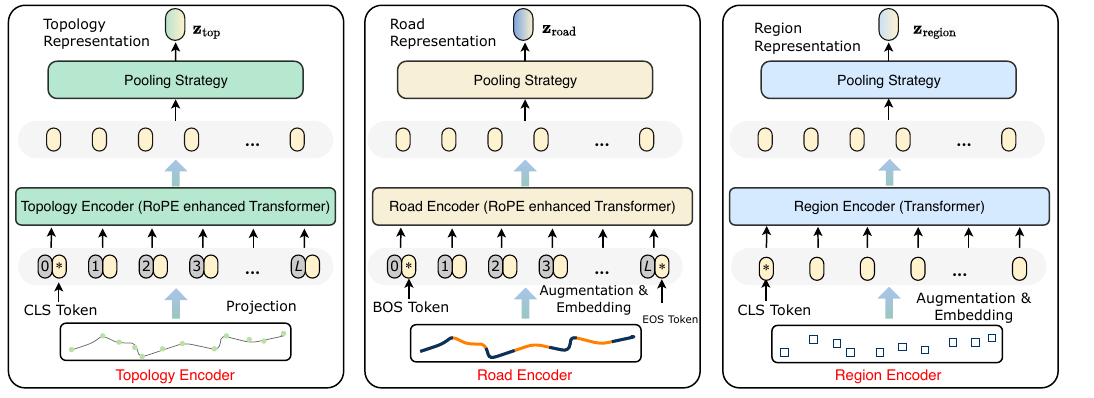}
    \Description[]{}
    \vspace{-3mm}
    \caption{The main structure of each encoder in the OmniTraj framework, the trajectory encoder is shown in the main content.}
    \label{fig:encoders_structure}
\end{figure*}
\section{Details of Framework}\label{app:structure}

\noindent \textbf{Structure of Encoders.}
As outlined in Section \ref{sec:method} and depicted in Figure \ref{fig:encoders_structure}, \model adopts a modular, decoupled design consisting of four independent encoders. 
Specifically, for the topology encoder, we first constructed a linear layer to project the topology points to a high-dimensional space and then added the [CLS] token and location information for processing by the RoPE-based Transformer block.
Finally, the acquired high-dimensional vector embeddings are processed using the corresponding pooling strategy to obtain a semantic representation of the topology.
For the road encoder, we replace the linear projection layer with an embedding layer to construct a unique embedding vector for each road identifier.
The rest of the process is the same as for the topological encoder.
For the region encoder, since we needed to cover as much of the sequence of the retrieval as possible, we removed the position encoding and used the native Transformer module.
Finally, for the projection head, which transforms individual modality-specific representations into a shared latent space, we use two stacked linear layers. This simple but effective approach ensures that the multimodal embeddings can be aligned and compared in a common space, facilitating flexible multimodal querying.

\begin{table}[h]
    \caption{Parameters setting and statistics of OmniTraj.}
    \vspace{-3mm}
    \centering
    \resizebox{1\linewidth}{!}{
    \begin{tabular}{lrrrr} 
    \toprule
     &  Traj. Enc & Top. Enc &  Road. Enc & Reg. Enc   \\ 
    \cmidrule(lr){1-5}
    Input dim  & 2  & 2  &  1 & 1   \\
    Output dim  & 256  &  256   & 256  & 256   \\
    Proj. dim  & 512  &  512   & 512  & 512   \\
    Max. Seq. length  & 200  & 128  &  128 &   64 \\
    Blocks  & 6  &  6 &  6 &  6  \\
    Pooling  &  cls & cls  & bos  &  cls  \\
    Atten. heads  &  8 & 8  &  8 &  8  \\
    Num. paras (M) & 4.75 & 4.75 & 4.17 & 4.81 \\

    \bottomrule
    \end{tabular}
    \label{tab:para_encoder}
    }
\end{table}

\noindent \textbf{Implementation Details}
We summarize the specific parameter settings for each module in OmniTraj, including dimensions, number of modules, and number of parameters.
Table \ref{tab:para_encoder} highlights the key configuration details for each of the encoders in \model. All encoders share the same output dimension of 256, and the projection dimension is set to 512 across all modalities. The encoders are designed with six Transformer blocks each, ensuring a balance between model depth and computational efficiency. The attention mechanism is equipped with eight attention heads to capture diverse semantic features. The number of parameters in each encoder is around 4.7 million, with minor variations depending on the modality. These settings ensure that \model is capable of handling large-scale trajectory data efficiently while maintaining flexibility and accuracy in multimodal retrieval tasks.

\section{Experiment and Setup}\label{app:exp_setup}
The experiments were conducted in a controlled environment with the following specifications. All models were trained on a machine equipped with an NVIDIA A100 GPU and evaluated with an A6000 GPU, which provided sufficient computational power for handling large-scale trajectory datasets. The code was implemented using Python 3.9 and PyTorch 1.8, leveraging the capabilities of CUDA for efficient GPU computation. 
We use the Adam optimizer InfoNCE loss with an initial learning rate of $2 \times 10^{-4}$.

\subsection{Dataset}\label{app:dataset}

We evaluated the model performance of OmniTraj and baseline on trajectory datasets from two cities, Chengdu and Xi'an. 
Following the approach in \cite{chang2023contrastive,chang2023trajectory}, we filtered out trajectories recorded outside urban areas or containing fewer than 20 points and transformed them to a fixed length of $200$ using
interpolation.
For the Chengdu dataset, there are 7597 road segments, and for Xi'an, there are 6018 identifiers. We divided the two cities into $16 \times 16 =256$ grids to represent the regions.
Table \ref{tab:dataset} summarizes the statistics of the resulting preprocessed datasets, where ``Avg. {}'', ``Avg. Road'', and ``Avg. Region'' indicates the average number of topological points, road segments, and regions per trajectory, respectively.

\begin{table}[h]
    \caption{Statistics of Two Trajectory Datasets.}
    \vspace{-3mm}
    \centering
    \resizebox{1\linewidth}{!}{
    \begin{tabular}{lcccc} 
    \toprule
    Dataset & \#Trajetory  & Avg. Topology &  Avg. Road &  Avg. Region   \\ 
    \cmidrule(lr){1-5}
    Chengdu  & \num{1224317}  & 13.33  &  21.08 & 9.52 \\
    Xian  & \num{1175949}  & 12.02 & 22.12 & 9.27 \\
    \bottomrule
    \end{tabular}
    \label{tab:dataset}
    }
    \vspace{-5mm}
\end{table}

\subsection{Baselines}\label{app:baseline}
For the trajectory similarity retrieval task, we use a range of heuristics and deep learning-based algorithms and compare variants of OmniTraj and its modality combinations.
Due to the high time complexity of heuristic methods, we restrict their evaluation to 1,000 query samples drawn from the 20,000 test trajectories, whereas the learning-based  approaches are evaluated on the entire testing set. 
\begin{itemize}[leftmargin=*] 

    \item \textbf{DTW} \cite{muller2007dynamic}. A classical method for computing the similarity between trajectories by optimally aligning their sequences using dynamic time warping. 
    
    \item \textbf{EDR} \cite{chen2005robust}:
    The Edit Distance on Real Sequence (EDR) metric measures trajectory similarity by considering spatial mismatches and is robust to noise and outliers. 

    \item \textbf{Hausdorff} \cite{alt2009computational}: A metric that quantifies the maximum deviation between two trajectories, measuring the worst-case distance between point sets.

    \item \textbf{Fréchet} \cite{alt1995computing}:  A distance measure that takes into account the continuity and order of points, often referred to as the “dog-walking” distance between curves. 
    
    \item \textbf{E2DTC} \cite{fang20212}: A self-supervised method that learns trajectory embeddings by capturing spatio-temporal patterns through clustering learning techniques.
    
    \item \textbf{t2vec} \cite{li2018deep}: A sequence-to-sequence deep learning-based approach that maps trajectories into a latent space using sequence modeling to capture the overall motion dynamics.

    \item \textbf{TrjSR} \cite{cao2021accurate}: Convert trajectories to images and use generative models combined with image super-resolution to learn trajectory representations.
    
    \item \textbf{TrajCL}: \cite{chang2023contrastive}: A recent contrastive learning-based framework specifically designed for trajectory similarity computation, which leverages contrastive learning for model training.

    \item \textbf{\model}: The \model framework utilizes only topology modality for optimal trajectory retrieval. 
    In addition, we tested a series of OmniTraj's variants, OmniTraj (modality), where modality represents different modes or their combinations. 
    For example, \model (road) leverages road segments to generate trajectory embeddings and \model (reg+road+top) combines region, road, and topology modalities.

\end{itemize}

For condition-based retrieval, since no existing work directly addresses this task, we adopt a set of simple approaches as baselines.
All of these baselines adopt InfoNCE for modality alignment.
\begin{itemize}[leftmargin=*] 
    \item \textbf{Embedding}:  It uses the embeddings generated from the lookup matrix, serving as a basic reference for comparison.

    \item \textbf{Linear}: A linear projection model where the trajectory embeddings are passed through a linear layer for further transformation.

    \item \textbf{CLIP}: CLIP encodes both the trajectory and condition (such as road or region) into shared space using vision and language principles (Bert), making it applicable for multimodal queries.

    \item \textbf{CLIP (LSTM)}: An variant of the CLIP model that uses an LSTM-based backbone to process sequential data instead of Bert.

    \item \textbf{CLIP (no-aug)}: This version of CLIP is evaluated without the use of data augmentation, providing a baseline to see the impact of augmentation techniques on retrieval performance.

    \item \textbf{OmniTraj}: The proposed method integrates modalities such as road segments, and regions for trajectory retrieval. This model is our primary approach for evaluating the effectiveness of multimodal retrieval in comparison to simpler models.

    \item \textbf{OmniTraj (no-aug)}: This baseline is similar to OmniTraj but does not utilize data augmentation during training. It serves as a control to assess how the augmentation impacts the model's performance on condition-based retrieval tasks.

\end{itemize}

\subsection{Evaluation Metrics}\label{app:metrics}

We employ multiple ranking metrics to evaluate the retrieval performance:
Mean Rank (MR) calculates the average position of the target sequence in the retrieved results. 
\begin{align}
\textbf{MR} = \frac{1}{|\mathcal{Q}|}\sum_{q=1}^{\mathcal{Q}} r_q,
\end{align}
A lower MR indicates better performance as the target appears closer to the top of the retrieval list.
Mean Reciprocal Rank (MRR) computes the average inverse rank of the first relevant result. 
Formally defined as:
\begin{align}
\textbf{MRR} = \frac{1}{|\mathcal{Q}|} \sum_{q=1}^{\mathcal{Q}} \frac{1}{r_q},
\end{align}
where $\mathcal{Q}$ is the set of queries, and $r_q$ is the rank position of the first relevant result for the $q$-th query. MRR ranges from 0 to 1, with higher values indicating better performance.
Hit Rate at k (HR@k) measures the percentage of queries where at least one relevant result appears in the top-k retrieved sequences. We report HR@1 and HR@5 to evaluate the model's ability to retrieve relevant sequences within the top-1 and top-5 results respectively.
\begin{align}
\textbf{HR@k} = \frac{1}{\mathcal{|Q|}} \sum_{q=1}^{\mathcal{Q}} \mathbb{I}(r_q \leq k),
\end{align}
where  $\mathbb{I}(\cdot)$  is the indicator function that returns 1 if the condition is satisfied, and 0 otherwise.

For condition-based retrieval, we evaluate the retrieval performance using Containment Rate (CR@k), which measures the proportion of query elements present in the top-K retrieved sequences. 
Specifically, CR@1 calculates the containment rate for the top-1 retrieved sequence, while CR@5 considers the union of elements from the top-5 retrieved sequences. 
Specifically,
we evaluate the retrieval results using Containment Rate (CR), which measures the proportion of query elements present in the retrieved sequence. The CR is calculated as:
\begin{align}
    \textbf{CR@k} = \frac{|\mathcal{Q} \ \cap \mathcal{R}_\textbf{k}|}{|\mathcal{Q}|},
\end{align}
where $\mathcal{Q}$ denotes the query set and $\mathcal{R}_\textbf{k}$ represents the union of elements from the top-k retrieved sequences. A higher \textbf{CR@K} indicates better coverage of the query elements in the retrieved results.

\section{Supplementary Experiments}
\begin{figure}[t]
    \includegraphics[width=1\linewidth]{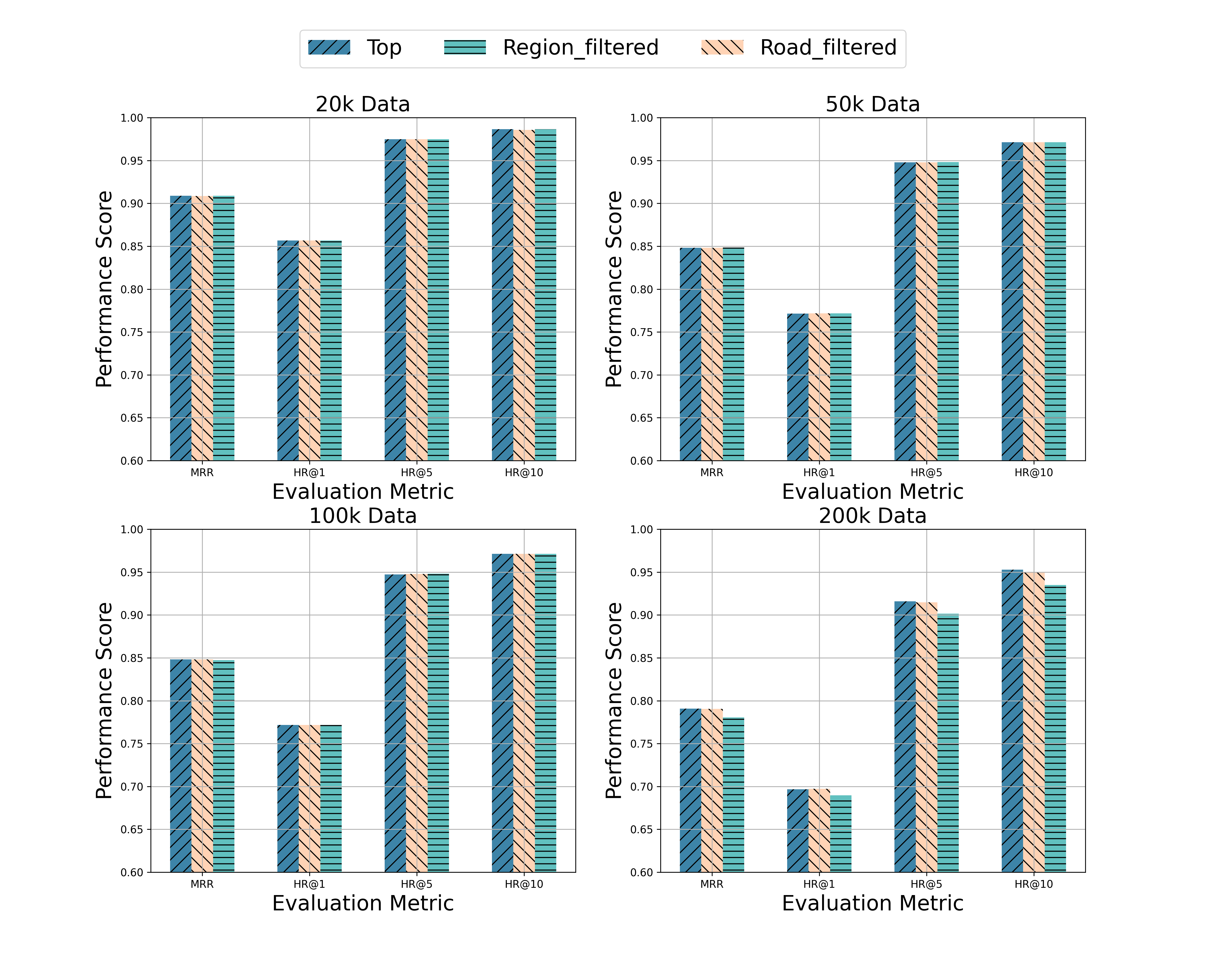}
    \vspace{-6mm}
    \Description[]{}
    \caption{Comparison of retrieval of trajectories using a two-stage approach with coarse-grained modal filtering followed by precise retrieval using topology modalities in a subset.}
    \label{fig:two_stage}
    \vspace{-3mm}
\end{figure}
\subsection{Two-Stage Retrieval}\label{app:exp_twostage}
We conduct our experiments using a two-stage approach for trajectory retrieval.
First, we employ coarse-grained modalities (such as roads and regions) to quickly filter and narrow down the dataset. Next, we apply topology-based fine-grained matching. We performed the experiments separately with different data sizes, ultimately reducing them to subsets of 200 for roads and 500 for regions.
As illustrated in \ref{fig:two_stage}, we observe no significant difference in subset retrieval after applying coarse-grained modal filtering compared to retrieval from the complete dataset. Notably, even as the data size increases to 200k, our method maintains strong robustness.
The above results show that our coarse-grained-based retrieval is highly flexible and is able to significantly reduce computational overhead and improve overall efficiency with this two-stage approach.

\begin{figure}[h]
    \includegraphics[width=1\linewidth]{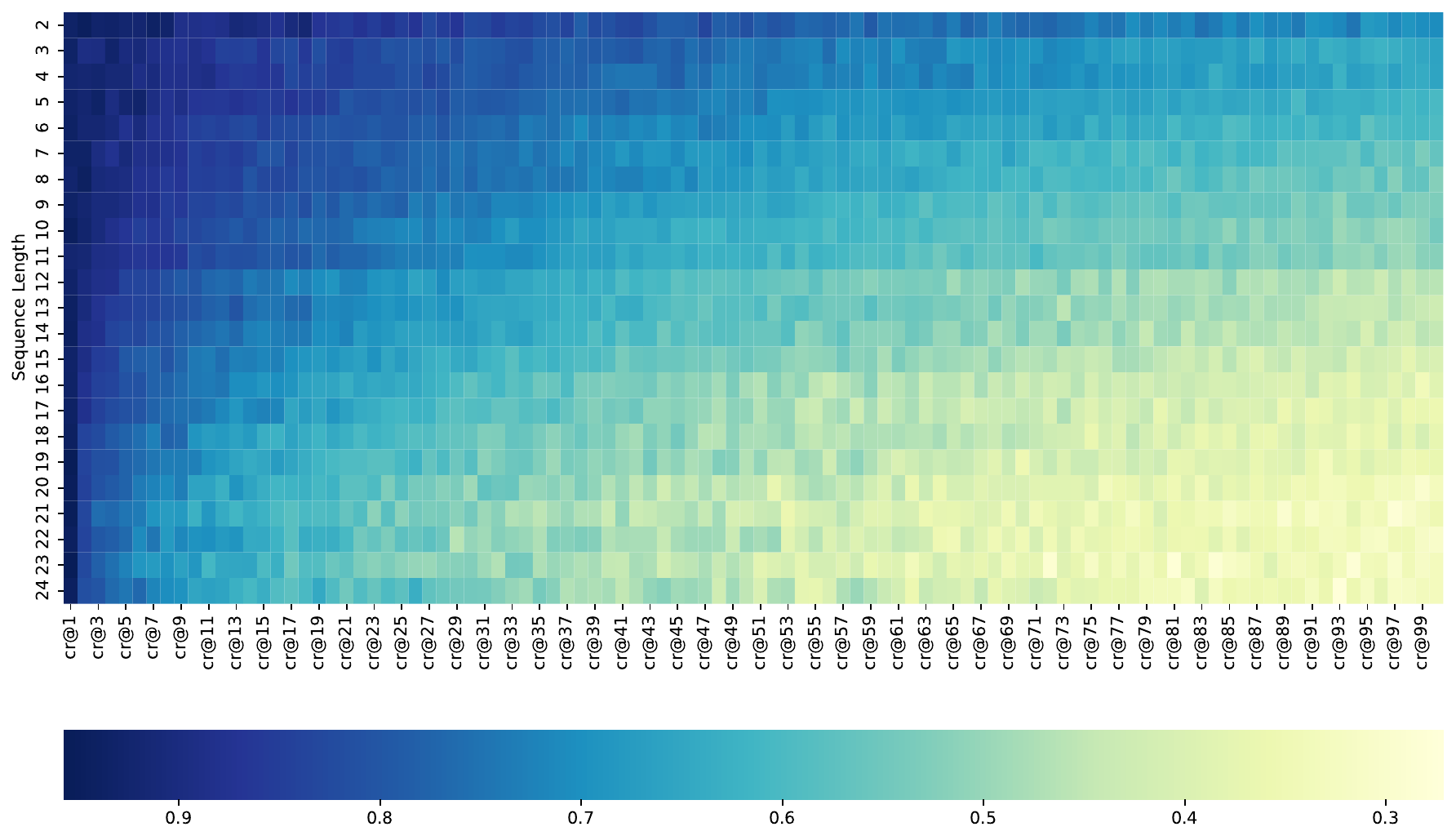}
    \Description[]{}
    \vspace{-6mm}
    \caption{Coverage changes with the length of the retrieved road sequence and the breadth of the retrieval. Sequence length increases from top to bottom and retrieval breadth increases from left to right.}
    \label{fig:coverage_topk}
    \vspace{-3mm}
\end{figure}

\subsection{Conditional Retrieval}\label{app:exp_condition}

Finally, we further explore the conditional retrieval flexibility and robustness of the \model.
The visualization results in Figure \ref{fig:coverage_topk} demonstrate two aspects: the first is the impact of the desired sequence length on the model, and the second is the range accuracy of the model retrieval.
It is clear from this result that the \model is sufficiently robust to the increase in retrieval length and does not sacrifice accuracy due to its increase.
Considering that our approach is not limited to exact sequence retrieval, it should also support queries such as retrieving trajectories that pass through or cover a specific sequence of road segments. 
We significantly increase the range (from 1 to 100) of the retrieval, and the results show that the sequences it retrieves can be well covered. Note that longer retrieved sequences imply more precise matches, and the fewer in the set to be retrieved satisfy this requirement, thus resulting in lower coverage.

\end{appendix}
\end{document}